# Prognosis and Treatment Prediction of Type-2 Diabetes Using Deep Neural Network and Machine Learning Classifiers

Md. Kowsher
*Dept. of Applied Mathematics*
Noakhali Science and Technology
University, Noakhali-3814,Bangladesh.
ga.kowsher@gmail.com

Mahbuba Yesmin Turaba
*Dept. of Information and Communication Technology*
Comilla University
Comilla, Bangladesh
mahbuba.yesmin11@gmail.com

Tanvir Sajed
*Dept. of Computing Science*
University of Alberta
Edmonton, Canada
tsajed@ualbarta.ca

M M Mahabubur Rahman
*Dept. of CSTE*
Noakhali Science and Technology
University Noakhali-3814, Bangladesh
toufikrahman098@gmail.com

*Abstract*—Type 2 Diabetes is a fast-growing, chronic metabolic disorder due to imbalanced insulin activity. As lots of people are suffering from it, access to proper treatment is necessary to control the problem. Most patients are unaware of health complexity, symptoms and risk factors before diabetes. The motion of this research is a comparative study of seven machine learning classifiers and an artificial neural network method to prognosticate the detection and treatment of diabetes with a high accuracy, in order to identify and treat diabetes patients at an early age. Our training and test dataset is an accumulation of 9483 diabetes patients' information. The training dataset is large enough to negate overfitting and provide for highly accurate test performance. We use performance measures such as accuracy and precision to find out the best algorithm deep ANN which outperforms with 95.14% accuracy among all other tested machine learning classifiers. We hope our high performing model can be used by hospitals to predict diabetes and drive research into more accurate prediction models.

## I. Introduction

Diabetes Mellitus (DM) is a very common metabolic disorder that affects millions of people worldwide. It occurs when the concentration of blood glucose reaches excessive level due to lack of production of insulin by the pancreas organ (Type 1 Diabetes) or due to insulin resistance (Type 2 Diabetes) [1]. It has been published that 422 million people are suffering from diabetes approximately in 2014 and it is expected to rise to 438 million in 2030[2, 3]. Among them, 90% of cases are Type 2 diabetes (T2DM) [4]. It may arise at an early childhood because of the failure of cells to respond to insulin appropriately [5]. So, patients have to face excessive tiredness, visual disorders, excessive thirst, skin infection recurrence, delayed wound healing and frequent discharge of urine [6]. It has been pointed out by Diabetes Research Center that 80 percent of cases of diabetes can be prevented or delayed if it is detected early [7]. Also, by controlling blood sugar, it is possible to lessen the T2DM effect. A healthy diet, physical exercise, sufficient nutrition for pregnant women, proper medication, weight at a necessary level are crucial to maintaining a safer sugar level.

When the diabetes is diagnosed with medical tests, it shows significantly dangerous symptoms but these methods do not perform well because of clinical complexity, time-consuming process and very high expense. However, using automated machine learning algorithms, a researcher can predict a disease like diabetes with reduced cost and time. In the field of Artificial Intelligence, classification is considered a supervised technique that analyses patient data and classifies whether or not the patient is suffering from a disease. Researchers have created different AI and machine learning techniques to automate prognosis of various diseases. Machine learning techniques studies algorithm and statistical model that has the capability for accurate prediction by using implicit programming. In medical science, they take the concept of the human brain as it contains millions of neurons to complete tasks of the human body. It is called nonlinear modelling and they are interconnected like brain cells although the neuron creation is done by program [8].

In this paper, first we have discussed various procedures and existing works about the prognosis of T2D , though we emphasized various classification algorithms known as Logistic Regression, KNN, Decision Tree, Naive Bayes, SVM, Linear Discriminant Analysis and Random forest classifier and Artificial Neural Network (ANN) for T2DM prediction. Our selected model is an Artificial Neural Network is found to be superior among all of them. Feedforward neural network contains the signal in one direction from the input to the output. It is used in different medical diagnostic applications such as nephritis disease, heart disease, myeloid leukemia etc. [ref].

We have taken a medical dataset from Noakhali Medical College, Bangladesh, consisting of 9483 samples and 14 symptoms per sample. The 80% data and 20% data are chosen to be training dataset and testing dataset respectively. Machine learning classification algorithms are applied to dataset and some elements may be missed. Then, the mean and median method is applied in order to detect it.

The contributions of this paper are summarized as

- We have proposed a prediction model for T2D using Artificial Neural Network machine learning classifier
- We have exerted seven classifier techniques and ANN on T2D data and provided comparison of accuracy among them.
- The improvement systems of the model, as well as accuracy, are mentioned in this work.

The remaining of the discussion is organized as follows: Section-II explains related work of various classification techniques for prediction of diabetes, Section-III describes the methodology and materials used, Section-IV discusses evaluated Results and Section-V delineates the conclusion of the research work.

## II. RELATED WORK

In recent years, several studies have been published using multiple machine learning classifiers, ANN techniques and various feature extraction methods. These have a drastic change in potential research and some works are discussed related to T2DM. Ebenezer et al. used the backpropagation feature of ANN in order to diagnose diabetes. It finds out the error by juxtaposing input and output number. Here, the preceding round error is greater than the present error each time by means of changing weight to minimize gradient of errors using a technique known as gradient descent [9]. Nongyao et al. delineated risk prediction by using various machine learning classification algorithms such as Decision Tree, Neural Network, Random Forest algorithms, Naïve Bayes, Logistic Regression. All of them followed Bagging and Boosting approaches to improve robustness except RFA [10]. Deepti et al. proposed a model to identify diabetes at a premature age by applying Decision Tree, SVM and Naïve Bayes on Pima Indians Diabetes Database (PIDD) datasets. They chose sufficient measures for accuracy including precision, ROC, F measure, Recall but Naïve Bayes beat them by acquiring the highest accuracy [11]. Su et al. applied decision tree, logistic regression, neural network,and rough sets to assess accuracy through various features like age, right thigh circumference, left thigh circumference, trunk volume and illustrates thigh circumference as a better feature than BMI in anthropometrical data [12]. Al-Rubeaan et al. has presented T2DM based on diabetic nephropathy (DP), then defined high impact risk factors; age and diabetes duration for microalbuminuria, macroalbuminuria and end-stage renal disease(ESRD) classifications[13]. Vijayan V. examines various types of preprocessing techniques which includes PCA and discretization. It increases the accuracy of Naïve Bayes classifier and Decision Tree algorithm but reduces SVM accuracy [14].

Micheal et al. proposed Multi-Layer Feed Forward Neural Networks (MLFNN) in order to diagnose diabetes by considering activation units, learning techniques on Pima Indian Diabetes (PID) data set and achieved 82.5% accuracy. It performs better than Naïve Bayes, Logistic Regression (LR) and Random Forest (RF) classifier [15]. Sadri et al. chose data mining algorithms like Naive Bayes, RBF Network, and J48 to diagnose T2DM for Pima Indians Diabetes Dataset that has 768 samples. Each sample has nine features as the total number of Pregnancy, Plasma Glucose Concentration, Diastolic Blood Pressure and 2-Hour Serum Insulin. Among them, the Naive Bayes algorithm is unbeatable and has 76.95% accuracy [16]. Pradhan et al. devised a classifier for diabetes detection using Genetic programming (GP) at low cost. Simplified function pool consists of arithmetic operations that are used in lower validation [17]. Yang Guo et al. applied Naïve Bayes classifier by using WEKA tool in order to predict Type2 diabetes and obtained remarkable accuracy [18].

Unlike these works, we have introduced diabetes's medication detection system using machining learning and deep ANN that will act like a doctor to choose the right medication of a patient suffering from diabetes.

## III. MATERIALS AND METHODS

In order to categorize diabetes therapy and drugs system for patients, the whole workflow is separated into four parts such as data collection, data preprocessing, training data via the proposed algorithms, and predictions. We have exerted seven machine learning classifiers and deep neural networks into the pre-processed data set.

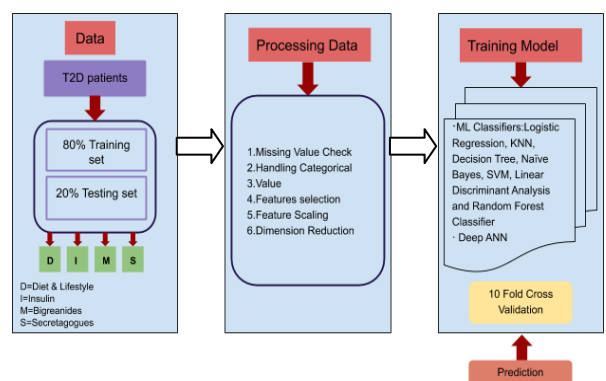

Fig.1. System Diagram of T2D analysis

The source of our data came from Noakhali Medical College, Bangladesh and the data set is separated into two parts such as training and test set. The training data are manipulated to the diagnostic system and 13 factors have been taken to determine therapy in order to apply machine learning and multilayer ANN. The dataset is tested from the trained machine learning classifiers and artificial neural network.

*A. Dataset*

As discussed before our data set contains information about 9483 diabetes patients and formatted in comma-separated-file (CSV). The dimension of the data set is 9483*14. It preserves 14 different kind of information of a diabetes patient such as 'Name of patient', "Fasting", "2 h after Pressure" "BMI", "Duration", "Age", "Sex", "Blood pressure", "High Cholesterols", "Heart Diseases", "Kidney Diseases", and , "Medications". The first 13 columns are considered independent variables and the last one is the dependent variable. It contains kinds of basic medicine name of diabetes such as Diet and Lifestyle Modification, Secretagogues, Biguanides, and Insulin.

When datasets consist of enough variables, it increases the accuracy of prediction. Here, "Fasting" measures blood test just before taking food, "2 h after glucose load" provides a blood test after two hours of eating. "BMI" refers to the weight and height of patients in kg/m². "Medication" indicates proper drugs and therapy. People who recovered T2DM at early stage follow some features: age group 30-75 years, diabetes of diagnosis duration is more than half years, glucose level at fasting plasma is higher than 125 mg/dl, creation of plasma indicates equal or greater than 1.7 mg/dl, plasma glucose after two hours is 11.

When a patient suffers from kidney problems, it may be a symptom of T2DM as higher sugar level may damage nephron. Even bleary eyesight is considered as a side effect for patients as eye's retina and the macula is affected. Bad cholesterol may lead to Diabetic dyslipidemia which can increase heart diseases and atherosclerosis. Here, we suggest treatment for kidney and v---ision problems. In order to categorize diabetes therapy and drugs system for patients, we applied seven machine learning classifiers and eight deep neural into a data system of Noakhali Medical College, Bangladesh. Training data are manipulated to diagnostic system and twelve factors have been taken to determine therapy in order to apply machine learning and multilayer ANN. For the training and testing of the systems, we divided the data set into 80% training and 20% test set. The training dataset is used to find out the appropriate model and best hyper-parameters and testing data set contains unseen data to predict the performance.

*B. Data preprocessing*

Data preprocessing involves raw data converting into a recognizable format from various sources. The well-preprocessed data aids for the best training of algorithms. Multi pre-processing training is held in our presented systems.

*1. Missing Value Check*

Usually, missing values may occur due to data incompleteness, missing field, programming error, manual data transfer from a database and so on. We may ignore missing values but it causes problems in parameter calculation and data accuracy for features such as age, wages and fare. We need to inspect whether a dataset has any missing value or not. There are many ways to handle missing values such as delete rows, missing values prediction, mean, median, mode and so on. But the most prominent policy for missing value replacement is the mean method and also it is used to exchange the approximate results in the dataset [19]. Mean is written in this way in mathematics,

$$\bar{x} = \frac{1}{n}\sum x_k \qquad (1)$$

Where, $\bar{x}$ denotes the mean and provides the average number of n.

*2. Handling Categorical value*

Categorical encoding identifies data type and transfers categorical features into numerical numbers as the majority of machine learning algorithms could not cope up with label data directly. Then numerical values are fed into the specific model. In our data set, there are five categorical variable names as 'Name of patients', "Heart Diseases", "Kidney Diseases", "Sex", and "Medications". There are two popular ways of transforming categorical data into numerical data such as Integer encoding and one-hot encoding. In the label encoder, categorical features are an integer value and contain a natural order relationship, but the multiclass relationship will provide different values for various classes. One hot encoding maps categorical value into binary vectors. Firstly, it is obvious to assign binary value to an integer value of female and male is 0 and 1. Then converting it to a 2 size of 2 possible integers in a binary vector. Here, a female is encoded as 0 and represented as [1, 0] in which index 0 has value 1 and vice versa. It chooses this value as a feature to influence model training [20].

*3. Features Selection*

Feature selection incorporates the identification and reduction of unnecessary features that have no impact on the objective function and high impact features are kept. Our dataset contains 14 types of elements and we have checked p-value which is a statistical process for finding out the probability for the null hypothesis. The features are taken out whose p-value indicates less than 0.05.

Moreover, multicollinearity refers to determine the high correlation which exists between two or more independent features and features that are influential to each other. It is called redundancy when two features are highly correlated. As we have to handle redundancy, it is essential to choose some methods such as χ2Test and Correlation Coefficient.

The Correlation Coefficient can be calculated by numerical data. Assume that A and B are two features and it can be defined as,

$$\sum_1^n \frac{(a_i - \bar{A}) + (b_i - \bar{B})}{n\sigma_a \sigma_b} \qquad (2)$$

After performing both p-value and multicollinearity test, we could come forward with seven features among thirteen independent features. Those are "Fasting", "2 Hours after Glucose Load" "Duration", "BMI", "High Cholesterols", "Heart Diseases", and "Kidney Diseases".

### 4. Feature scaling

Most of the time, the dataset does not remain on the same scale or even not normalized. So, feature scaling is a fundamental data transformation method for coping the dataset to algorithms. We need to scale value of features and provide equal weight to all features in order to obtain the same scale for all data. Moreover, it is possible for scaling to change in different values for different features. There are lots of techniques for feature scaling for example Standardization, Mean Normalization, Min-Max Scaling, Unit Vector and so on.

In our research work, we have taken Min-Max Scaling or normalization process as the features are confined within a bounded area. Minmax normalization is a z-series normalization to transform linearly x to x' where maxX and minX are the maximum and minimum value for X respectively.

$$x' = \frac{x - min(x)}{max(x) - min(x)} \qquad (3)$$

When x=max, then y =1 and x=min, y=1.

The scaling range belongs between 0 and 1(positive value) and -1 to 1(negative) and we have taken the value between 0 and 1.

### 5. Dimension Reducing

Dimensionality reduction refers to minimizing random variables by considering the principal set of variables that avoids overfitting. For a large number of dataset, we need to use dimension reduction technique. In our study, we prefer dimension reduction for dimensional graphical visualization. There are a lot of methods for reducing dimension, for instance, LDA, PCA, SVD, NMF, etc. In our system, we have applied Principal Component Analysis (PCA). It is a linear transformation based on the correlation between features in order to identify patterns. High dimensional data are estimated into equal or lower dimensions through maximum variance. We have taken two components of PCA according to their high variance so that we can graphically visualize in Cartesian coordinate system.

### C. Training Algorithms

The training dataset for T2DM is applied to each algorithm to find out medications and model performance is assessed by obtaining accuracy.

#### a. Machine Learning Classifier

Since we focus on the performance of treatment predictions, we have implemented seven machine learning classifiers such as logistic regression, KNN, SVM, Naive Bayes, decision tree, LDA, random forest tree.

Logistic regression is based on the probability model; it is derived from linear regression that mapped the dataset into two categories by considering existing data. At first, features are mapped linearly that are transferred to a sigmoid function layer for prediction. It shows the relationship between the dependent and independent values but output limits the prediction range on [0, 1]. As we need to predict the right treatment of a diabetes person, it is beneficial to use a binary classification problem.

Linear Discriminant Analysis (LDA) belongs to a linear classifier to find out the linear correlation between elements in order to support binary and multiclass classification. The chance of inserting a new dataset into every class is detected by LDA. Then, the class that contains the dataset is detected as output. It can calculate the mean function for each class and it is estimated by vectors for finding group variance.

Support Vector Machine (SVM) is the most recognized classifier to make decision boundary as hyperplane to keep the widest distance from both sides of points. This hyperplane refers to separating data into two groups in two-dimensional space. It performs better with non-linear classification by the kernel function. It is capable of separating and classifying unsupported data.

K-nearest neighbours (KNN) works instant learning algorithm and input labeled data that act as training instance. Then, the output produces a group of data. When k=1, 2, 5 then it means the class has 1, 2 or 5 neighbours of every data point. For this system, we choose k=5 that means 5 neighbours for every data point. We have taken Minkowski distance to provide distance between two points in N-dimensional vector space to run data. Suppose, points p1(x1, y1) and p2(x2, y2) illustrates Minkowski distance as,

$$d_\mathbf{p} : (x, y) \mapsto \|x - y\|_p = \left( \sum_{i=1}^n |x_i - y_i|^p \right)^{\frac{1}{p}} \qquad (4)$$

Here, d denotes Minkowski distance between p1 and p2 point.

Naive Bayes Classifier is constructed from Bayes theorem, in which features are independent of each other in present class and classification that counts the total number of observations by calculating the probability to create a predictive model in the fastest time. It outperformed with a huge dataset of categorical variables. The main benefits of that it involves limited training data to estimate better results. Naive Bayes theorem probability can be derived from P (T), P(X) and P (X|T). Therefore,

$$P(X) = \frac{(P(T)P(T))}{P(X)} \qquad (5)$$

The decision tree is a decision-supporting a predictive model based on tree structure by putting logic to interpret features. It provides a conditional control system and marks red or green for died or alive leaves. It has three types of nodes: root node, decision nodes and leaf nodes. The root node is the topmost node among them and data are split into choices to find out the decision's result. Decision nodes basically comprise of decision rules to produce the output by considering all information gain and oval shape is used to denote it. The terminal node represents the action that needs to be taken after getting the outcome of all decisions.

Multiple random trees lead to the random forest to calculate elements of molecular structure. A decision tree looks like a tree that is the storehouse of results from the random forest algorithm and bagging is applied to it in order to reduce bias-variance trade-off. It can perform feature selection directly and output represents the mode of all classes. In Random Forest Tree, we took the total number of trees in the forest: 10.

b. *Artificial Neural Network*

An ANN is considered as a human brain due to consisting millions of neurons to communicate with each other. It has three layers; the input layer fed raw data to network, hidden layer is the middle layer based on input, weight and the relationship denoted by activity function. Output layers value is determined by activity, weight and relationship from the second layer.

Since we need to find out the probability of each treatment and the objective function is not binary, so we used softmax activation function instead of sigmoid between the hidden layer and output layer. There is no rule of thumb to choose hidden layer in ANN. If our data is linearly separable then we don't need any hidden layer. Then the average node between the input and output node is preferable.

In our system, we prefer six hidden layers between the input node and the hidden layer and 25 epochs to train a neural network. It has no gradient vanishing problem and uses ReLU activation function to train dataset without pretraining.

c. *Validation*

The validation is a technique of evaluating the performance of algorithms. It cooperates to evaluate the model and reduce overfitting. Different types of validation method includes Holdout method, K-Fold Cross-Validation, Stratified K-Fold Cross-Validation and Leave-P-Out Cross-Validation. We have picked out k-fold validation dataset is divided into k subsets in k times. One k subset act as test set and error is estimated by average k trails. Therefore, k-1 subsets produce training set. We prefer k=10 generally which contains 10 folds, repeat one time and stratified sampling as each fold has a similar amount of samples.

IV. EXPERIMENTAL RESULT ANALYSIS

A. *Experimental tool*

The whole task has been implemented in python 3.6 programming language in Anaconda distribution. Python library offers various facilities to implement machine learning and deep learning. The unbeatable library for data representation is pandas that provide huge commands and large data management. We have used it to read and analyze data in less writing. Afterward, scikit-learn has features for various classification, clustering algorithms to build models. Also, Keras combines the advantages of theano and TensorFlow to train a neural network model. We use to fit and evaluate function to train and assess neural network model respectively bypassing the same input and output, then we apply matplotlib for graphical visualization.

B. *Model performance*

For boosting performance, it is always a better idea to increase data size instead of depending on prediction and weak correlations. Also, adding a hidden layer may increase accuracy and speed due to its tendency to make a training dataset overfit. But partially it is dependent on the complexity of the model. Contrarily, increasing the epochs number ameliorate performance though it sometimes overfits training data. It works well for the deep network than shallow network when considering regulation factor.

Hereafter, we have added another hidden layer; choose epoch 100 then the Deep ANN accuracy risen up to 95.14% which is superior among all of them.

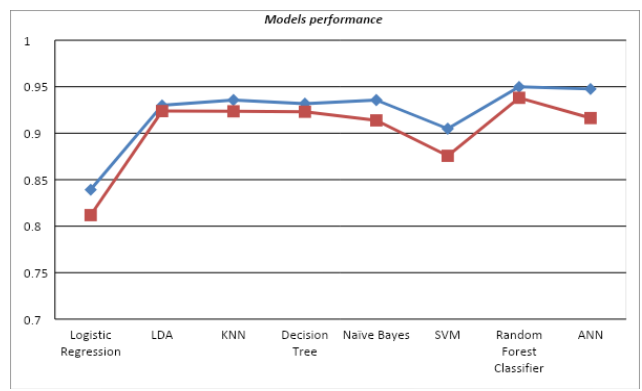

Fig.2. Models Performance Comparison.

C. *Improving Model performance*

For boosting performance, it is always a better idea to increase data size instead of depending on prediction and weak correlations. Also, adding a hidden layer may increase training accuracy and speed due to its tendency to make training dataset overfit. But partially it is dependent on the complexity of the model. Contrarily, increasing the epochs number ameliorate performance though it sometimes overfits training data. It works well for the deep network than shallow network when considering regulation factor.

Hereafter, we added another hidden layer; choose epoch 100 then the Deep ANN accuracy of the training and test set is risen up to 96.42% and 95.14% which is superior among all of them.

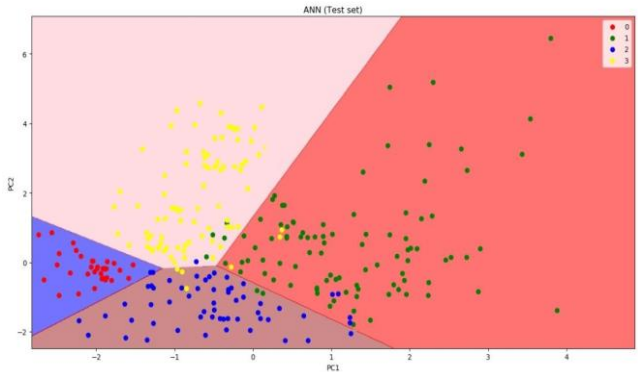

Fig.3. 2-D Graphical Visualization of Test set

### D. Final Result

After applying feature extraction to the dataset and implementing several types of classification and deep neural network, we found artificial neural network as better performer with best validity and Random forest classifier are preferable among other machine learning classifiers.

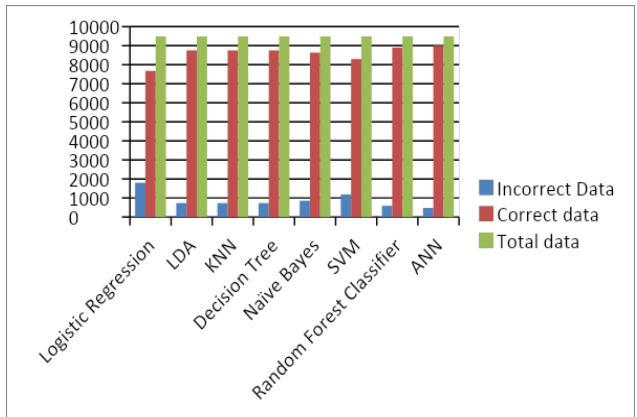

Fig.4. Final Result Comparison

## V. CONCLUSIONS

Type-2 diabetes can lead to a lot of complications as heart attack, kidney damage, blurred vision, hearing problems and Alzheimer's disease. The main problem is lower accuracy of the prediction model, small datasets and inadaptability to various datasets. In this paper, the medication and treatment are predicted by a comparative study of seven machine learning algorithms and deep neural networks. Artificial neural networks play a vital role in medical science by minimizing classification error that leads to greater accuracy. Experiment result determines the designed ANN system achieved higher accuracy of 94.7%. It can cooperate with experts to detect T2DM patients at a very early age and provide the best treatment option.

In the future, we can enhance the accuracy of early treatment to lessen the suffering of patients. Also, we can implement more classifiers to pick up the leading one for record-breaking performance and extend it to automation analysis. There is a plan to apply this designed system in diabetes or for other diseases. It may increase the performance of prediction of various diseases. Larger dataset leads to the higher training set and it cooperates in advanced accuracy. It is convenient for people to have an application on their smartphones related to T2DM that may have T2DM symptoms, treatment, risk factors, and health management.